\documentclass[10pt,twocolumn,letterpaper]{article}

\usepackage{wacv}
\usepackage{times,epsfig,graphicx,amsmath,amssymb,xcolor}
\usepackage{bm,comment}
\usepackage{color,subfigure}
\usepackage{soul}
\usepackage[normalem]{ulem}
\DeclareMathOperator*{\argmin}{arg\,min}
\def\etal{{et al. }}
\def\lsd-slam{{LSD-SLAM }}

\wacvfinalcopy 

\def\wacvPaperID{155} 

\ifwacvfinal\fi
\begin{document}

\title{Computing Egomotion with Local Loop Closures for Egocentric Videos}

\author{Suvam Patra \thanks{Equal Contribution} \\
IIT Delhi
\and
Himanshu Aggarwal $^\ast$ $^\dagger$ \\
Qualcomm Inc.
\and
Himani Arora $^\ast$ \thanks{Research assistant at IIIT Delhi during this project} \\
Columbia University
\and 
Subhashis Banerjee \\
IIT Delhi
\and
Chetan Arora \\
IIIT Delhi
}

\maketitle
\ifwacvfinal\thispagestyle{empty}\fi
\begin{abstract}
Finding the camera pose is an important step in many egocentric video applications. It has been widely reported that, state of the art SLAM algorithms fail on egocentric videos \cite{ego-ff, hyperlapse, ego-seg, activity-rec}. In this paper, we propose a robust method for camera pose estimation, designed specifically for egocentric videos. In an egocentric video, the camera views the same scene point multiple times as the wearer's head sweeps back and forth. We use this specific motion profile to perform short loop closures aligned with wearer's footsteps. For egocentric videos, depth estimation is usually noisy. In an important departure, we use 2D computations for rotation averaging which do not rely upon depth estimates. The two modification results in much more stable algorithm as is evident from our experiments on various egocentric video datasets for different egocentric applications. The proposed algorithm resolves a long standing problem in egocentric vision and unlocks new usage scenarios for future applications.
\end{abstract}

\section{Introduction}

\begin{figure}[t]
\centering
\includegraphics[width=0.99\linewidth]{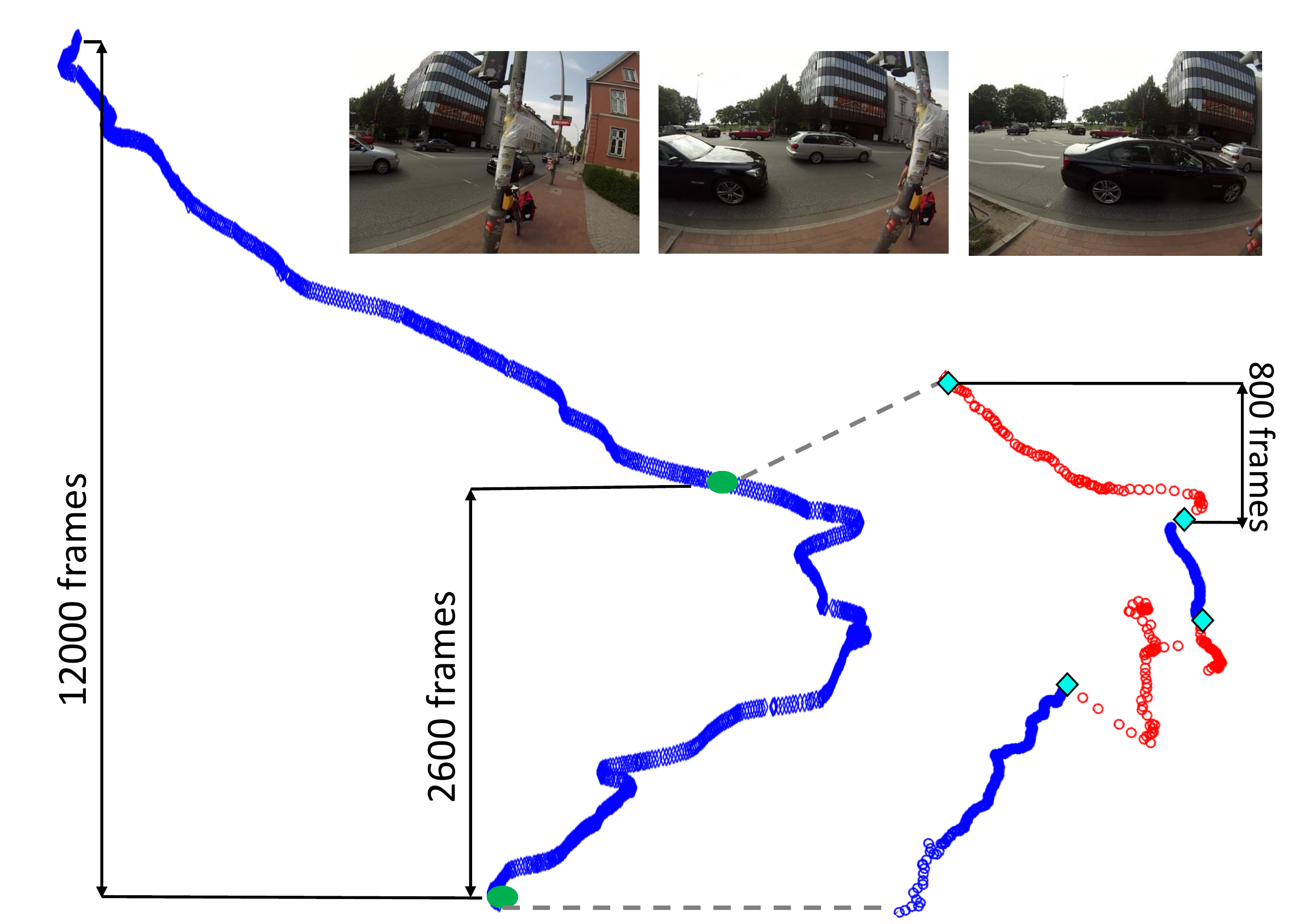}
\caption{We present an egomotion estimation technique targeted for egocentric videos. The figure on the right shows a trajectory computed from \lsd-slam \cite{lsd-slam} on a sequence from \cite{hyperlapse}. The different color segments show breaks in the trajectory computation every few 100 frames. On the other hand our algorithm runs successfully even after 12000 frames (on the left). The proposed method resolves a long standing problem in egocentric vision.}
\label{fig:intro}
\end{figure}

Ability to record hands-free videos led to initial popularity of wearable cameras like Google glass \cite{google-glass} and GoPro \cite{gopro}. These wearable cameras are typically harnessed to the wearer's head, and provide a first person (wearer's) view of the surroundings. The captured videos are therefore also referred to as egocentric videos. With the improvement in design and usability of egocentric cameras, such cameras are now being used for life logging, law enforcement, assistive vision as well as smart home applications.

Owing to their unique first person perspective and always-on nature, egocentric videos are rich in content usually unseen from point and shoot videos. This has generated lots of interest among the computer vision community. Computer vision researchers are mining egocentric videos for understanding wearer's activities \cite{ego_ac_recog_1, ego_ac_recog_2, ego_ac_recog_3, ego_ac_regog_5_lowres, jpl}, gaze fixation \cite{ego-seg}, temporal segmentation \cite{ego-seg}, video summarization \cite{Grauman-Story:CVPR13, ego_ac_regco_and_chaptering, Grauman-Important:CVPR2012} etc. In many of these applications, computing the pose of the camera and hence the position and gaze of the wearer is important.

For a long time, the egocentric community has struggled with the problem of egomotion estimation \cite{ego-ff, hyperlapse, ego-seg, activity-rec}. Though visual odometry algorithms \cite{lsd-slam, rgbd-vo-icra, semi-dense-vo-iccv-13, ptam, dtam} have been quite successful for handheld or vehicle mounted cameras, the  relatively `wild' nature of egocentric videos: unrestrained camera motion, wide variety of indoor and outdoor scenes as well as moving objects, makes it extremely challenging for the egomotion estimation algorithms. Head rotation of the wearer causes the scene to change rapidly making any kind of feature tracking in egocentric video a non-trivial task. Any error in tracking causes the feature based visual odometry methods to lose putative matches, and ultimately lead to their failure. Even when the tracking is not lost, sharp rotations introduce inaccuracies in pose estimates, and, in the absence of any recovery mechanism, the estimations drift to  points of instability creating breaks in the trajectory. The typical motion profile of linear forward velocity due to wearer walking/driving in egocentric videos, kills any standard loop closure opportunity relying on similar camera pose.

An alternate approach is to employ the more robust dense or semi-dense SLAM \footnote{Because of their overlap, we will be using SLAM/Odometry techniques interchangeably in our discussions, though our focus is only on the egomotion estimation and not on the associated 3D map.} \cite{lsd-slam,rgbd-vo-icra,semi-dense-vo-iccv-13} techniques which utilize a greater portion of the image. Even though these methods perform better than their sparse counterparts, they are still unable to handle the wild dynamics of egocentric videos.

It is to be noted that shaky handheld videos do not have motion profiles similar to egocentric videos. For example, even a fast moving hand-held camera typically do not have dominant 3D rotation and usually scans a scene point multiple times giving standard loop closure opportunities.

In this paper we present a technique for camera egomotion estimation targeted especially for egocentric videos.  We use these to correct pose inaccuracies using rotation averaging \cite{govinduefficient} and Gauss-Newton re-initialization.  The specific contributions of the paper are as follows:
\begin{enumerate}
    \item We present a fast and robust egomotion estimation method for egocentric videos.
    \item We suggest a novel local loop closure technique aligned with the wearer's head motion for egocentric videos. We exploit the typical to-and-fro sweeps in egocentric camera motion to detect local loop closures. While in all previous work for handheld videos, loop closures are typically carried out at the end, the proposed local loop closures aligned with the head sweep of the wearer enable fixing of estimation errors locally and efficiently. This not only makes the  algorithm  stable but also improves the accuracy significantly. The suggested improvement is specific to egocentric videos only.
    \item While standard loop closure techniques typically fix both rotation and translation, the methodology fails in the presence of noisy depth estimates common in egocentric setting. We suggest to fix only rotation using 2D techniques which is both faster and more stable. We show the improvement in computed trajectory by fixing rotation alone.
    \item We experiment on multiple problems such as EgoSampling \cite{ego-ff}, Hyperlapse \cite{hyperlapse}, Temporal Segmentation \cite{ego-seg}, Activity Recognition \cite{activity-rec} and Gaze Fixation \cite{ego-seg}. For each such problem the state-of-the-art egomotion estimation has been reported to fail \cite{ego-ff,hyperlapse,ego-seg,activity-rec}. However the proposed method works flawlessly on all of these, indicating the robustness and efficacy of the technique.
\end{enumerate}

\begin{figure*}[t]
\centering
\subfigure[]{\includegraphics[width=0.24\linewidth]{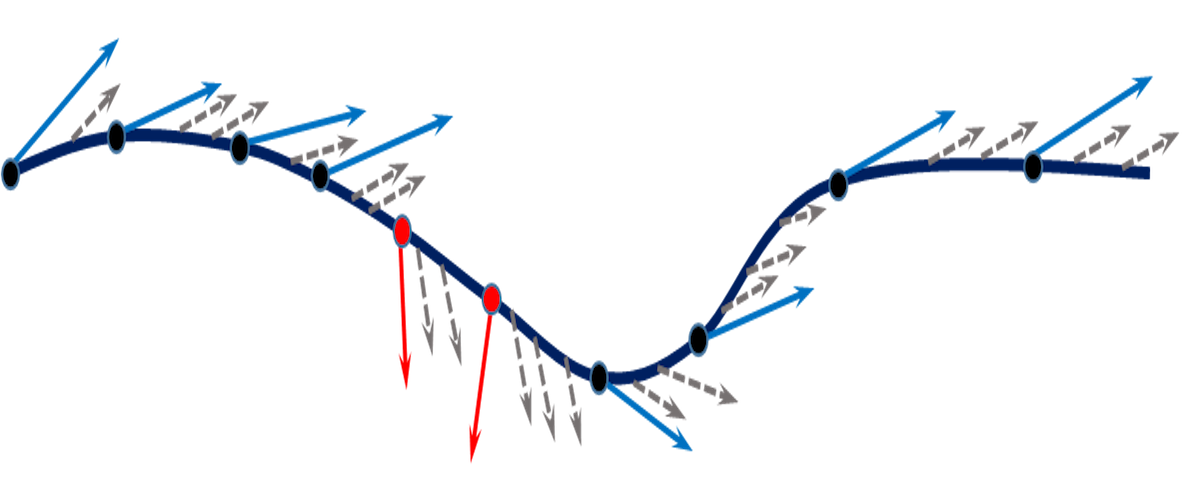} }
\subfigure[]{\includegraphics[width=0.24\linewidth]{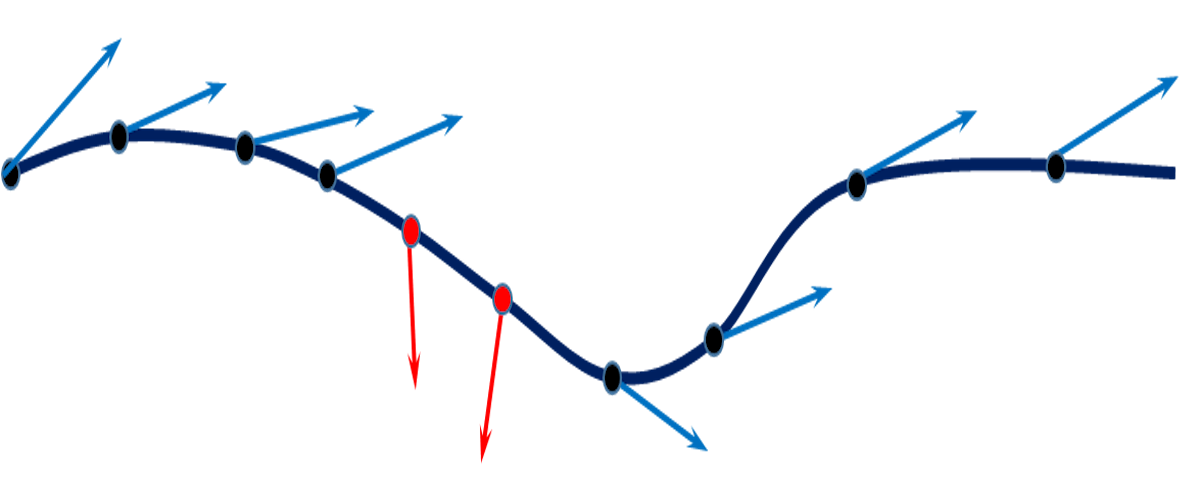} }
\subfigure[]{\includegraphics[width=0.24\linewidth]{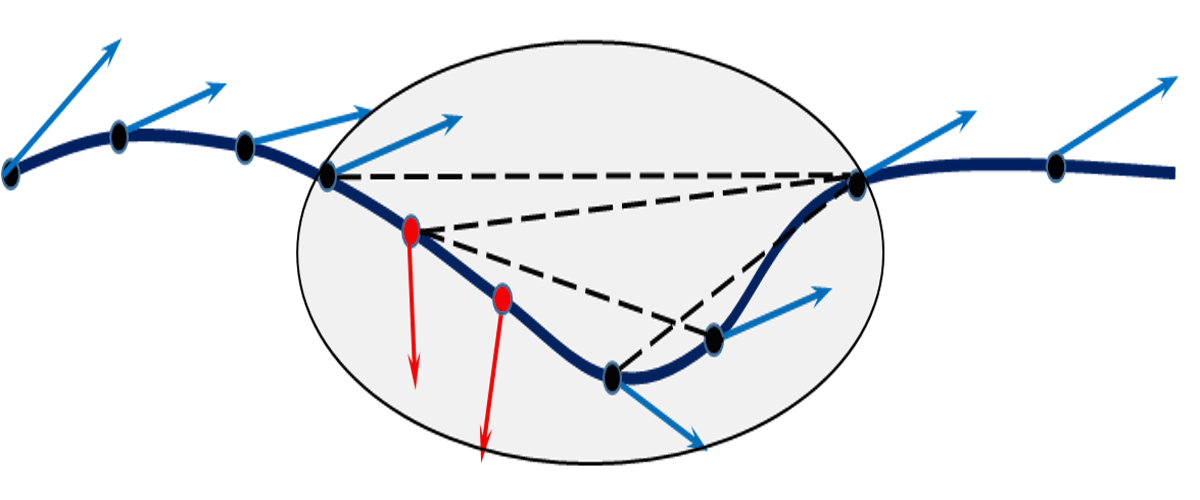} }
\subfigure[]{\includegraphics[width=0.24\linewidth]{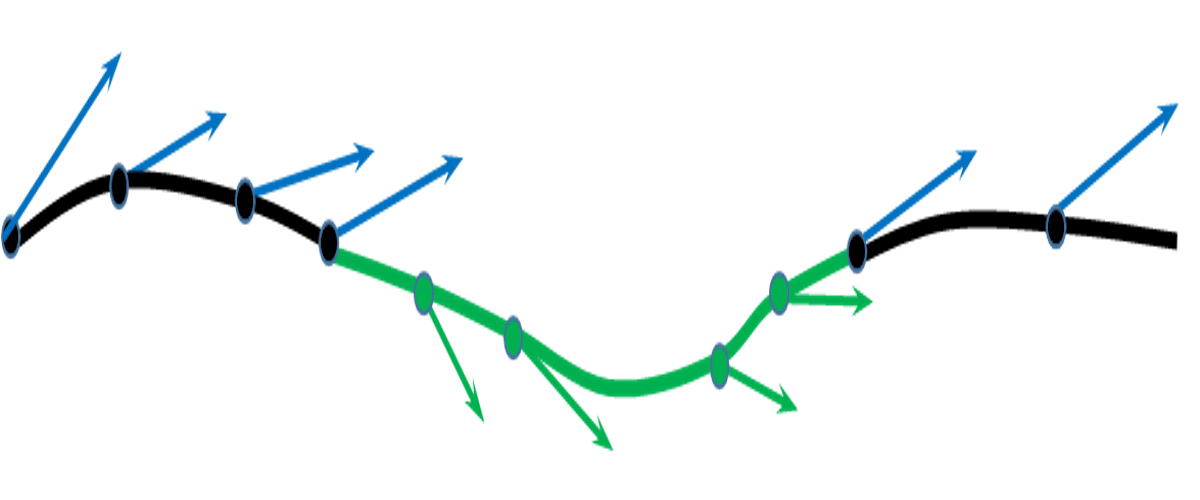} }
\caption{Our approach to correction of poses using short loop closures and rotation averaging. (a) shows a typical trajectory where each camera is connected to the next, and due to large rotation two cameras (indicated in red) are wrongly estimated. The dotted arrows represent intermediate frames (b) only keyframes take part in loop closures (c) shows short loop closure detection using KL divergence and extra matches (dotted edges) are found  to close the loops  (d) shows the corrected trajectory (green) after running rotation averaging and one iteration of Gauss-Newton for correcting camera positions. Images are best viewed in colour.}
\label{Fig:posecorrection}
\end{figure*}

Fig. \ref{Fig:posecorrection} gives a schematic representation of the proposed approach.

\section{Related work}

Visual odometry algorithms can be classified into two broad categories: stereo and monocular. Stereo based visual odometry algorithms can usually give information about the absolute scale of the world. However, when the distance to the scene is much larger compared to the stereo baseline, these methods give no real advantage and hence, monocular visual odometry algorithms are typically used. Monocular methods employ only a single camera and suffer from the problem of scale ambiguity. The scale may be resolved by the integration of sensors such as IMU or with scene calibration objects.

Further classification of these algorithms can be done on the basis of the techniques used for pose estimation, namely feature based methods, dense methods and hybrid methods that employ a combination of both. Feature-based methods mainly consist of two steps \cite{ptam,orb-slam}. In the first step, important features are extracted from the images and matched. In the second step, the camera poses are estimated using only these sparse feature points. The reduction in the number of points used in the estimation process greatly reduces the computational load thereby increasing the speed. However, a large part of useful information in the scene is ignored.

Dense methods use the entire image and not just few selected features \cite{dtam}. The camera poses are estimated as the set of parameters which minimize the image difference over all pixels in the image. To increase the accuracy of estimation, semi-dense methods are usually adopted, which perform photometric error minimization only in regions of sufficient gradient \cite{lsd-slam,semi-dense-vo-iccv-13}.

The work closest to ours is \lsd-slam \cite{lsd-slam} which does dense tracking directly on $sim(3)$, to explicitly detect scale-drifts. \lsd-slam builds upon \cite{semi-dense-vo-iccv-13}, to continuously estimate a semi-dense inverse depth map for the current frame, which in turn is used to track the motion of the camera using dense image alignment. Given an inverse depth map, both methods estimate the camera motion using non-linear minimization in combination with a coarse-to-fine scheme, as originally proposed in \cite{rgbd-vo-icra}. The minimization is done using weighted Gauss-Newton optimization on Lie-Manifolds. Our method also uses a similar optimization technique for the initial camera pose estimation.

As discussed in detail in \cite{Williams2009}, loop closures are detected using three major approaches in literature - map-to-map, image-to-image and image-to-map. L. Clemente \etal \cite{clemente_etal_rss2007}  use a map-to-map approach where they find correspondences between common features in two sub-maps. M. Cummins and P. Newman \cite{CumminsIJRR08} use visual features for image-to-image loop-closures.  Matching is performed based on presence or absence of these features from a visual vocabulary. B. Williams \etal \cite{WilliamsIROS08} use an image-to-map approach and find loop-closure using re-localization of camera by estimating the pose relative to map correspondences. Using the assumption that aerial video views a roughly planar ground and homographies can be used to register the frames, Leotta \etal \cite{homo} have proposed a homography-guided loop closure algorithm to address the long-term loop closure problem. In this paper, we use an image-to-image approach for the purpose of detecting loop closures.

\begin{figure*}[t]
\centering
\includegraphics[width=0.99\textwidth]{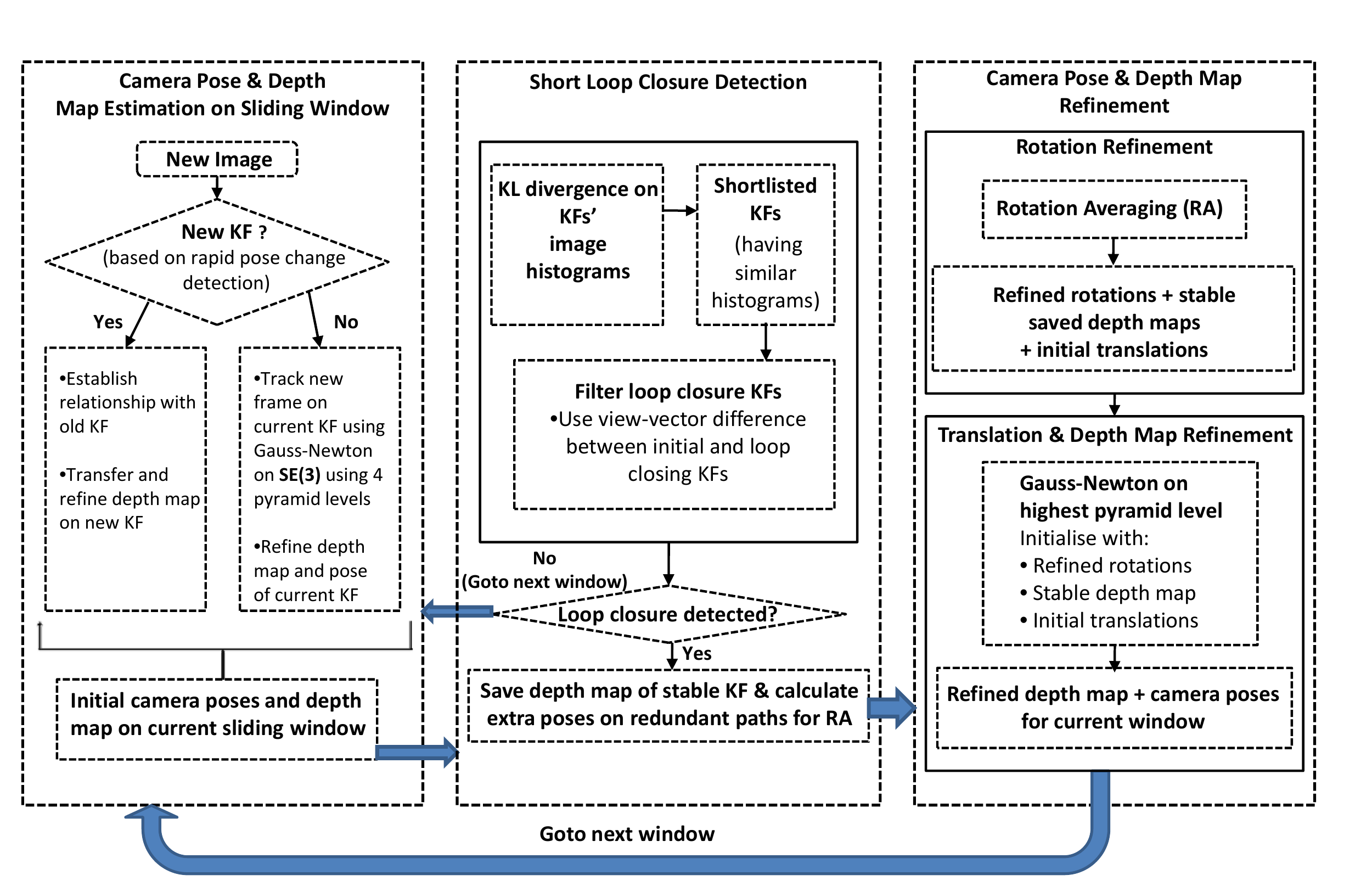}
\caption{Our framework. (KF here denotes keyframes)}
\label{fig:frameworkmain}
\end{figure*}

\section{Background}

The 6 DOF camera pose of an image $I'$ w.r.t a reference image $I$ refers to the rotation and translation of its camera centre.  The camera pose is represented as an element of a Lie Group by a $\mathrm{4 \times 4}$ matrix $\mathbf{g} \in \mathrm{SE(3)}$ which comprises of a $\mathrm{3 \times 3}$ rotation matrix $\mathrm{\mathbf{R} \in SO(3)}$ and a $\mathrm{3 \times 1}$ translation vector $\mathbf{t}$: $\mathbf{g} = \begin{bmatrix} \mathbf{R} & \mathbf{t} \\ \mathbf{0} & \mathbf{1} \end{bmatrix}$. In Lie Algebra, the camera pose is represented by a 6 element vector $\bm{\xi} \in \mathfrak{se}(3)$: $\bm{\xi} = \begin{bmatrix} \mathbf{w} & \mathbf{v} \end{bmatrix}$, where, $\mathrm{\mathbf{w}}$ and $\mathrm{\mathbf{v}}$ represent the rotation and translation respectively.

We estimate the camera pose by aligning an image $I'$ with a reference image $I$ which minimizes the photometric error between the two image frames. Following the notation given in \cite{rgbd-vo-icra}, the problem can be reduced to one of minimizing the sum $\sum_{i \in I} r_i$, where, the residual $r_i$, at a pixel $i$ is given as: $r_{i} = I'(\tau (\bm{\xi}, \mathbf{x_{i}})) - I(\mathbf{x_{i}})$. Here $\bm{\xi}$ is the estimated camera pose of $I'$ w.r.t $I$ and $\tau: \mathbf{x} \rightarrow \mathbf{x'}$ is the warping function mapping a pixel in $I$ to a pixel in $I'$.

The warping function can be constructed as follows. Using the inverse projection function, $\pi$, the 3D world point corresponding to a pixel $\mathbf{x}$ is given as: $\mathbf{X}=\pi^{-1}(\mathbf{x}, Z(\mathbf{x}))$, where $Z(x)$ is the depth at a pixel $\mathbf{x}$. When $\mathbf{X}$ is viewed in the coordinate frame of the second camera, it is transformed by the pose $\mathbf{g}=\exp(\bm{\xi})$ of $I'$ w.r.t $I$ to get the 3D world point $\mathbf{X'}$ given by $\mathbf{X'}= T(\bm{\xi}, \mathbf{X}) = \mathbf{R} \mathbf{X} + \mathbf{t}$. When $\mathbf{X'}$ is projected on the second camera, we get the warped pixel coordinates $\mathbf{x'} = \pi(\mathbf{X'})$.

\subsection{Pose Estimation}

The camera pose $\xi$ can be computed as the one minimizing the residual: $\argmin_{\bm{\xi}} \sum_{x_{i}}w_{i}(r_{i}(\bm{\xi}))^{2}$. The residual is calculated over all pixels $x_{i}$ having  valid depth estimates. The weight $w_i$  at a pixel indicates the confidence measure of the estimated depth. The objective function is non-linear (since $\tau$ is non-linear in $\bm{\xi}$) and can be solved using Gauss-Newton optimization. We assume that the current estimate of $\bm{\xi}$ is known and iteratively refine the estimate as $\bm{\xi} \leftarrow \bm{\xi} + \Delta\bm{\xi}$, where $\Delta\bm{\xi}$ is computed as:
\begin{multline}
 \underset{\Delta\bm{\xi}}{\argmin}\sum_\mathbf{x} w_{i}[I'(\tau(\mathbf{x};\bm{\xi}+\Delta\bm{\xi}))-I(\mathbf{x})]^{2} = \\  \sum_\mathbf{x} w_{i}[\underbrace {I'(\tau(\mathbf{x};\bm{\xi})) + \nabla I' \frac{\partial \tau}{\partial \bm{\xi}} \Delta \bm{\xi}}_{\text{First order Taylor approximation}}-I(\mathbf{x})]^{2}
\end{multline}
where, $ \nabla I'$ is the image gradient $ \begin{bmatrix}
\frac{\partial I'}{\partial x} & \frac{\partial I'}{\partial y}
\end{bmatrix}$ and
$\frac{\partial \tau}{\partial \bm{\xi} }$ is the Jacobian of the warping function.

The pose of each image frame $I'$ is calculated with respect to a keyframe $I$. The poses thus obtained are concatenated to estimate the pose with respect to the first keyframe (assumed to be the world origin with $\bm{\xi}=0$). Since only the depth corresponding to the reference frame $I$ is required in estimating the warping function,  only a keyframe is associated with a depth map. The keyframes are switched at regular intervals to maintain sufficient overlap between the frames.

\subsection{Depth Map Estimation}

The previous section assumed that the depth at a pixel $Z(\mathbf{x})$ is known apriori. We jointly estimate both the camera pose and the scene depth by computing stereo matches along  epipolar lines at different values of depth. The SSD error is calculated at each of these pixels and the one with minimum error is chosen to be the correct match. Since initially both the depth and camera pose is unknown, the algorithm is bootstrapped with a random depth map having high weight only at those pixels where sufficient gradient is available. After a few keyframe propagation, the depth map converges to a reasonable approximation.

\subsection{Rotation Averaging} \label{sec:rot}

Whenever the gaze direction returns to a neutral position following a deviation to the left or to the right, we detect local loop closures using a procedure outlined later in Section~\ref{sec:loopclosures}. The loop closures provide multiple redundant rotation estimates from the current frame to past frames - both through pair-wise estimates, and direct rotation estimates with a few archived key frames. The multiple redundant rotation estimates along different paths allow us to  average out the rotations to obtain consistent rotations along all sub-paths. We denote by $\mathfrak{C}$ the set of cameras in a window selected after loop-closure detection.

Let us assume that the relative rotation between camera $i$ and camera $j$ be denoted by ${\bf R}_{ij}$ and the absolute rotations be denoted by $\{{\bf R}_1,\cdots,{\bf R}_N\}$ (where $N$ is the number of cameras in the current window). Then the following relationship should hold:
\begin{equation}
{\bf R}_j{\bf R}_i^{-1} = {\bf R}_{ij} \label{Eqn:RelativeRotation}
\end{equation}
We estimate the absolute rotations $\{{\bf R}_1,\cdots,{\bf R}_N\}$ by minimizing the sum of errors between the estimated pairwise rotations ${\bf R}_{ij}$ and the relative rotations calculated as ${\bf R}_j{\bf R}_i^{-1}$ from equation \ref{Eqn:RelativeRotation} as:
\begin{equation}
\{{\bf R}_1,\cdots,{\bf R}_N\}=\displaystyle \argmin_{\{{\bf R}_1,\cdots,{\bf R}_N\}}\sum_{(i,j) \in \mathfrak{C}} d^2\left({\bf R}_j{\bf R}_i^{-1}, {\bf R}_{ij}\right)\label{Eqn:RotationAveraging}
\end{equation}
where
\begin{equation}
d({\bf R}_1,{\bf R}_2) = \frac{1}{\sqrt{2}}||\log({\bf R}_2 {\bf R}_1^{-1})||_F
\label{frobenius}
\end{equation}
which is the intrinsic bi-variate distance measure defined on the manifold of 3D rotations or the Special Orthogonal group $SO$(3) and $||.||_F$ is the Frobenius norm. We use the robust function and the  technique described in~\cite{govinduefficient} to solve the optimization problem defined in Equation~\ref{Eqn:RotationAveraging}.

\section{Proposed Algorithm}

Our overall framework is depicted in Figure~\ref{fig:frameworkmain}. In what follows we describe some of the key steps.

\subsection{Short local loop closures} \label{sec:loopclosures}

For a typical hand held video, global loop closure is an important step to fix the accumulated errors over individual pose estimation. In egocentric setting, where the motion of the wearer is linear forward, a user may not revisit a particular scene point, which makes global loop closure impossible. Furthermore, given the usual noisy measurements due to wild nature of egocentric videos, the computed trajectories tend to drift quickly unless fixed by loop closures. We note that, a wearer's head typically scans the scene left to right and back during the course of natural walking. The camera looks at same scene multiple times, thus forming a series of short local visual loop closures. We take advantage of this phenomenon to improve the accuracy of the estimated camera poses.

We maintain an archive of the past keyframes, which is continuously updated whenever a new keyframe is initialized. An incoming frame $I'$ is not only mapped to the current keyframe $I$ but also to the previous keyframes $I_{t-1}, I_{t-2} \cdots$ for establishing redundant paths. Initially, when a new frame is received, an attempt is made to find a match with an archived keyframe. A match is said to exist when the KL Divergence \cite{KLDiv}, computed between the histograms of two query images, is small and the estimated camera view vectors point to the same direction. This actually puts a constraint on the current frame that it is looking in the same direction or view as of the keyframe under consideration. This constitutes a local loop closure. In our experiments, we select a key frame for every 10 frames. To keep the loop closures local, only the neighbouring keyframes are searched up to a fixed time interval $T$. Additionally the loop closures are detected on sliding windows for imposing a smoothness constraint over the trajectory.

\begin{figure}[t]
\centering
\includegraphics[width=0.99\linewidth]{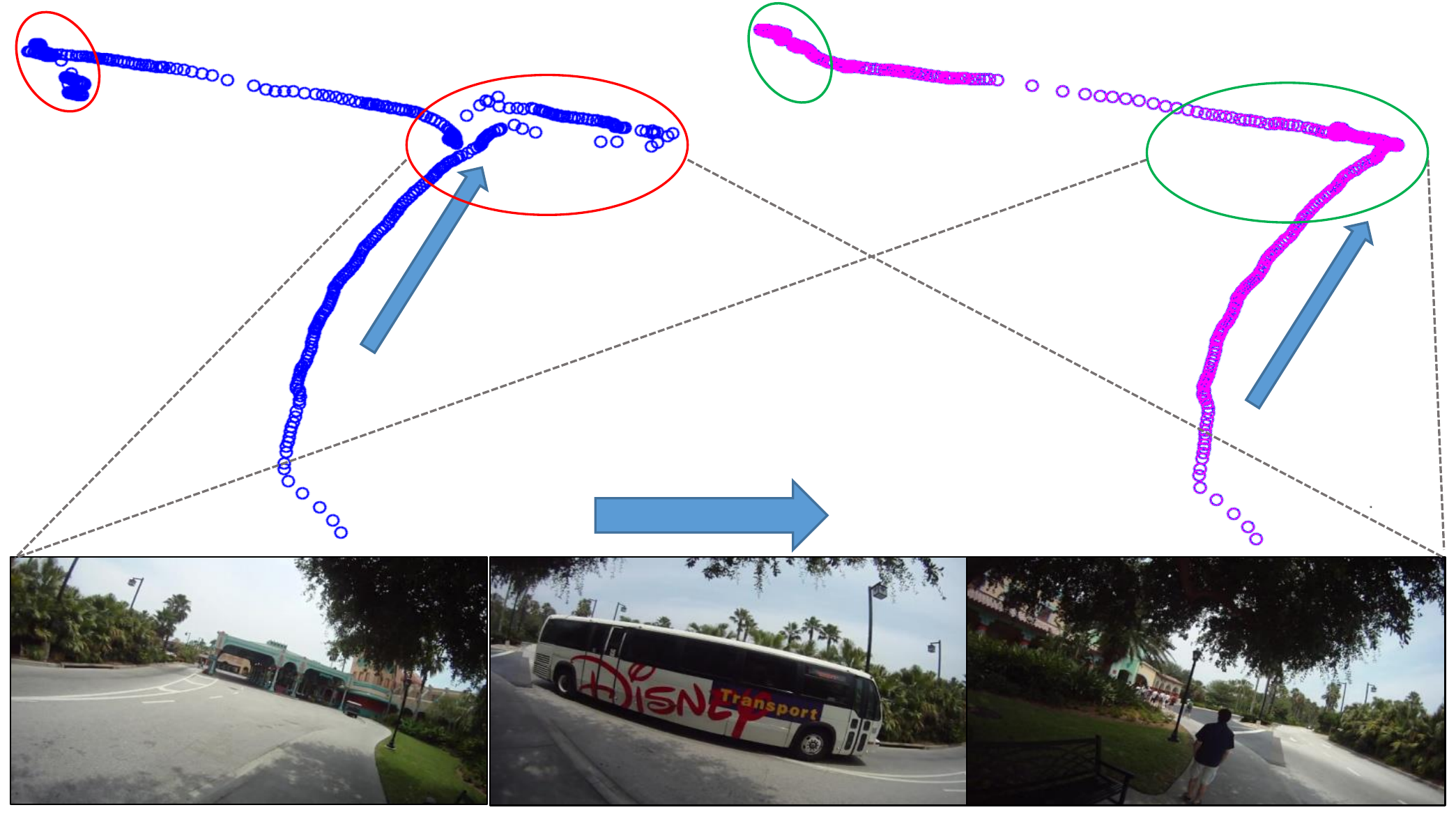}
\caption{Effect of rotation averaging and Gauss-Newton re-initialization is shown on the Georgia Tech. Social Interaction data set \cite{social-interaction}.  Trajectory in blue is before refinement with red ovals depicting breaks in the trajectory due to large rotations (some sample images are also shown), which gets corrected after refinement by our method in pink. The trajectory is depicted for a sequence of 500 frames on which we detected  multiple batches of short local loop closures within a time frame of 5-10 seconds. Images are best viewed in colour.}
\label{fig:rotavg}
\end{figure}

\subsection{Rotation averaging based pose refinement}

Short local loop closures give rise to a number of redundant paths between the keyframes. These additional constraints are then used to refine the initial pose estimate. Sharp rotations over a longer period cause the rotation estimates to drift. This drift is corrected by rotation averaging (see section \ref{sec:rot} for more details). Rotation averaging is performed in small sized windows to ensure enough flexibility for errors to be averaged out. This continuous cycle of pose estimation followed by rotation averaging  over a short loop closure ensures that consistent and accurate estimations of the prior camera pose are used to initialize the next set of incoming frames. Given the tracking challenges and noisy 3D estimates from an egocentric video, the main source of error in egocentric camera pose estimation is the reliance on these noisy 3D estimates. We do a rotation stabilization using 2D, image based, pairwise rotation estimates, which is not effected by such noisy 3d estimates.

\subsection{Gauss-Newton re-initialization}

Although rotation averaging gives robust estimates of  camera rotations, it does not alter the camera translations. However, the new rotation estimates are likely to change  the camera center estimates. The new camera centers are then re-estimated by running a second level of Gauss-Newton  optimization using the newly averaged rotations and previously calculated translations as initial estimates. A single iteration at the highest resolution of the image-pyramid ensures that the rotation estimates remain nearly unchanged and only minor readjustments are brought about in the translations, thereby keeping the final pose estimates restricted in the local neighborhood/minima. This can be achieved at minimal computational cost. Moreover, this approach is more suitable than simply fixing rotation and updating the translation estimates without 3D refinement, which may make the estimates incompatible and erroneous.

Fig.\ref{fig:rotavg} shows a result from the proposed depicting advantage of short loop closure followed by rotation averaging.

\section{Experiments}

\begin{figure*}[t]
\centering
\subfigure{\includegraphics[width=0.241\linewidth]{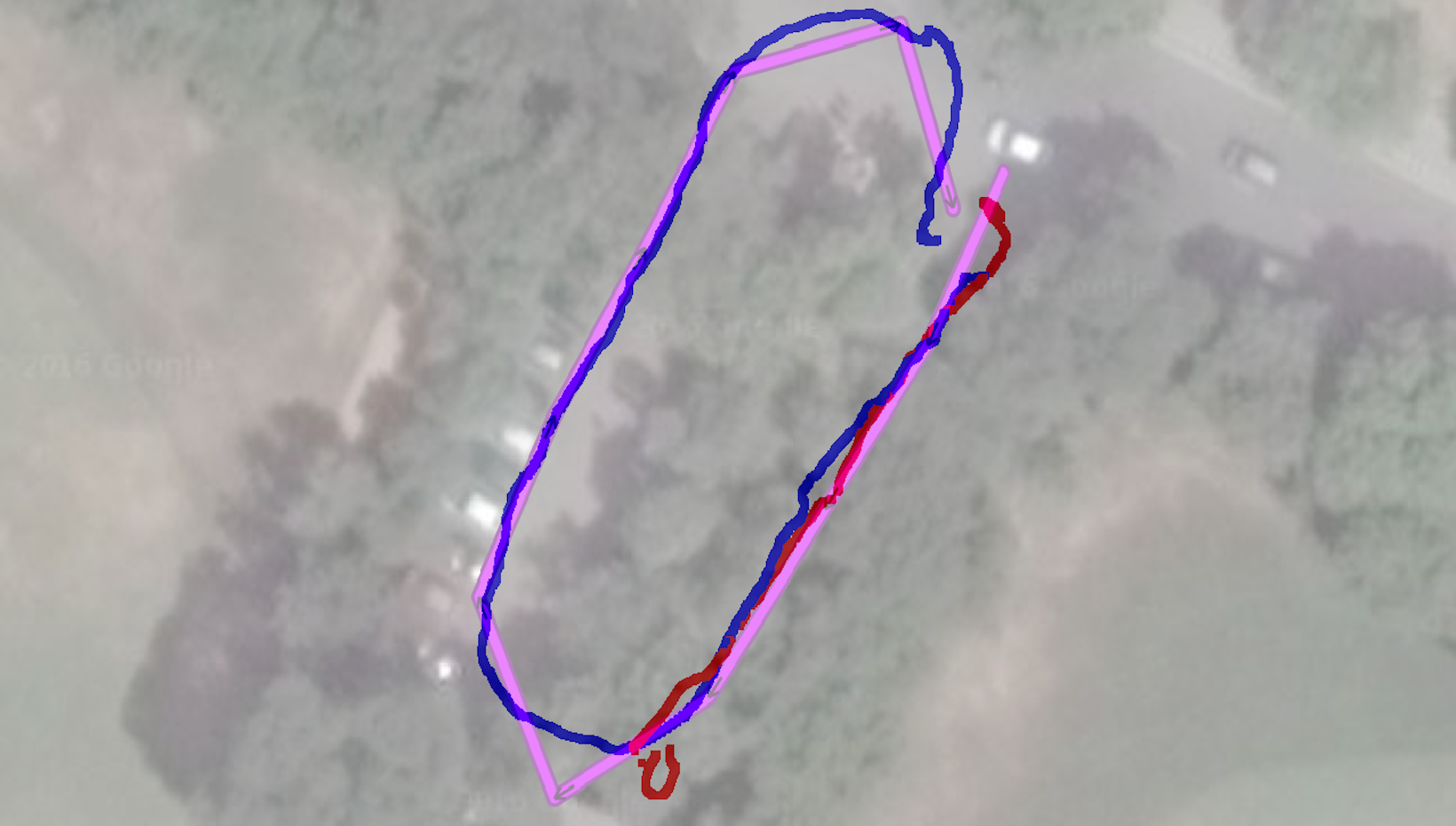} }
\subfigure{\includegraphics[width=0.241\linewidth]{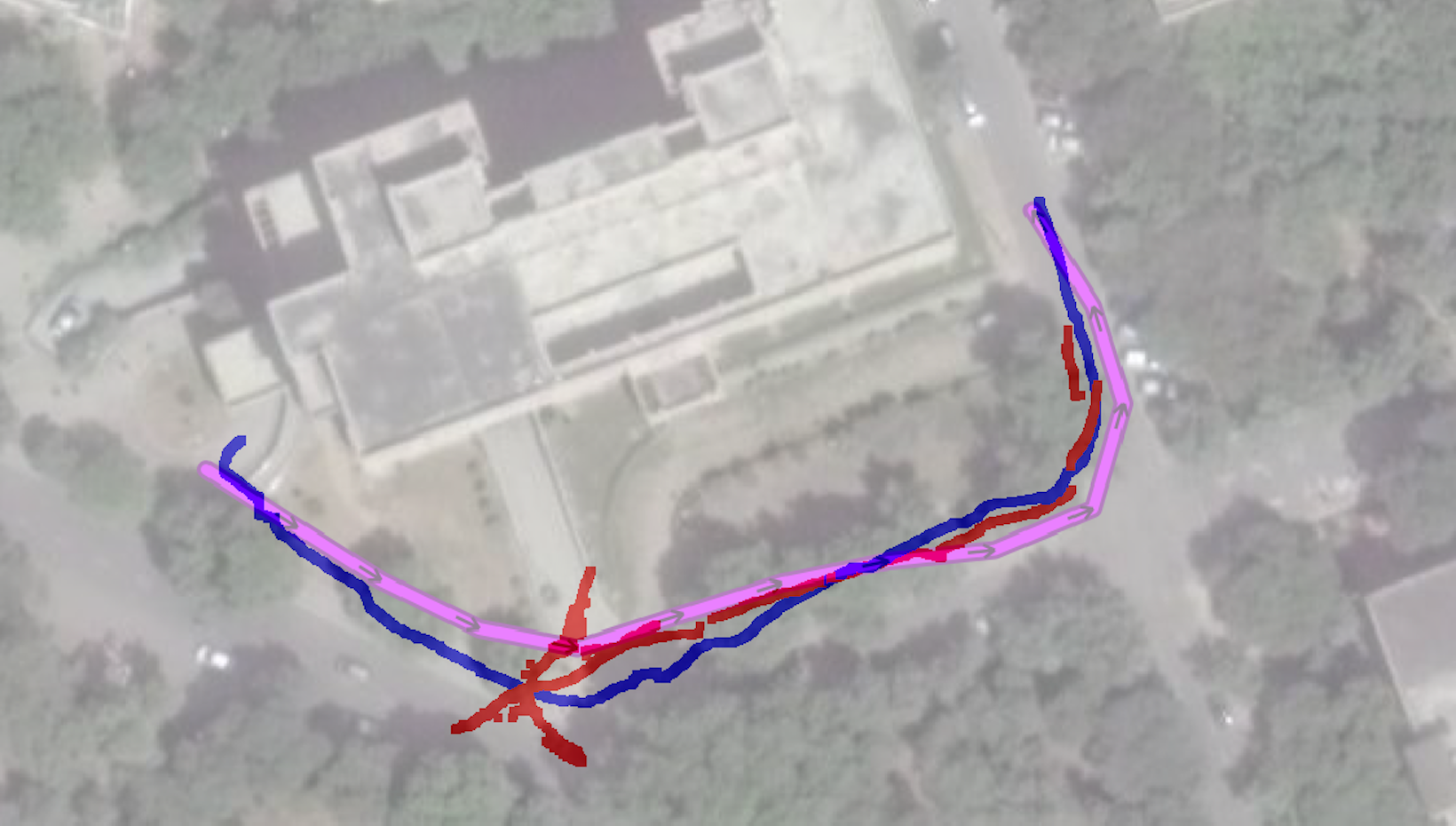} }
\subfigure{\includegraphics[width=0.241\linewidth]{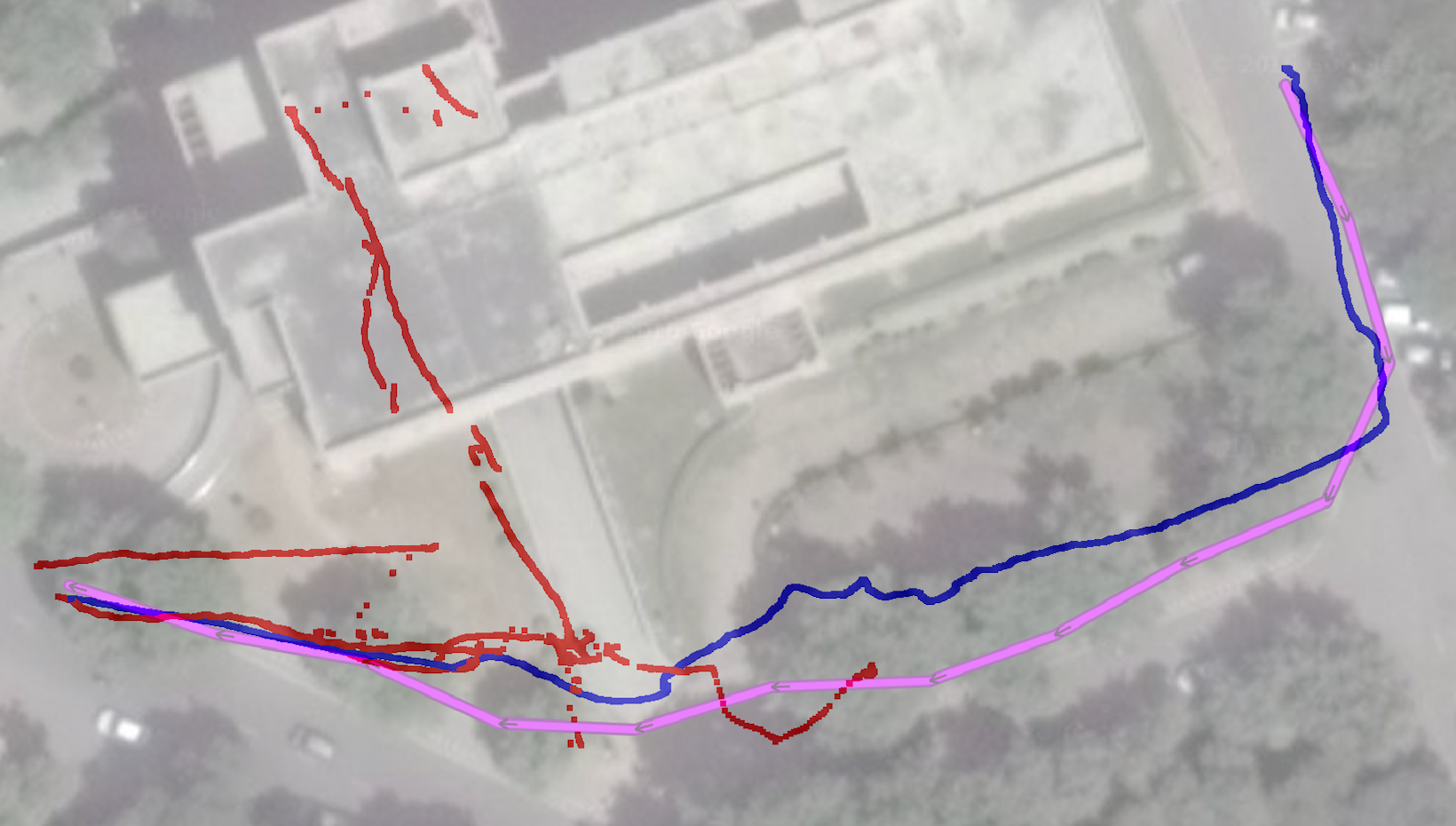} }
\subfigure{\includegraphics[width=0.242\linewidth]{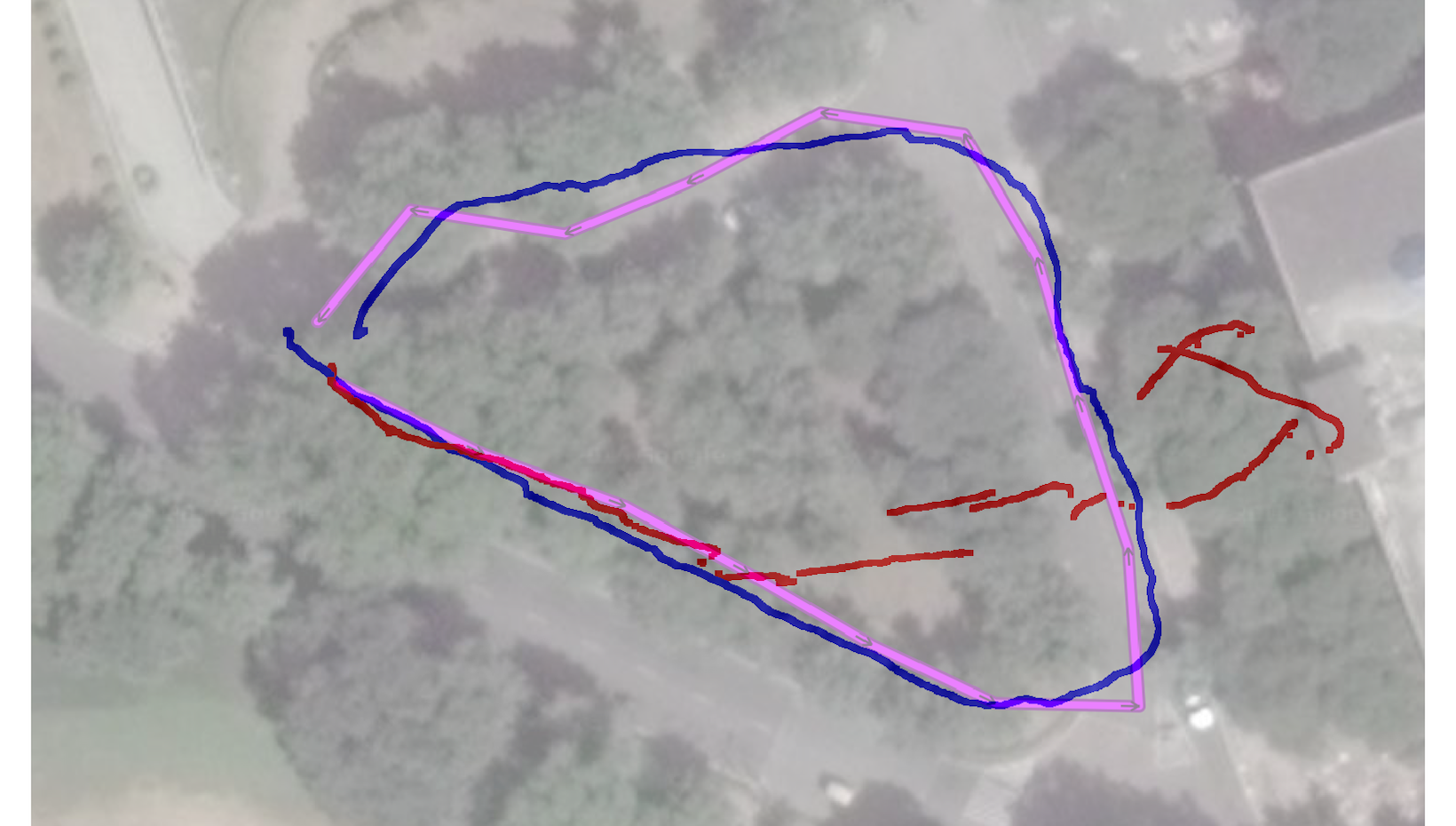} }
\caption{Comparison of trajectories obtained for egocentric videos. In each of these images the trajectories obtained by our method are marked in blue, while those attained by LSD-SLAM are marked in red. Due to the lack of any standardized datasets with ground truth poses for egocentric videos, the accuracy of the trajectories is demonstrated by overlaying them on GPS data (pink). It should be noted that the LSD-SLAM trajectories shown here are the ones obtained just before the algorithm crashed. The frequent crossed trajectories in LSD-SLAM results is due to noisy 3D estimates maintained by the algorithm which often leads to false matches}
\label{Fig:trajectories}
\end{figure*}

Robust and accurate camera pose estimation can effectively be used in numerous egocentric applications. However, state-of-the-art visual SLAM solutions such as \cite{lsd-slam} fail to find accurate trajectories for egocentric videos. This has led to various researchers adapting alternate, sub-optimal approaches in egocentric video based problems \cite{ego-ff, ego-seg, hyperlapse, activity-rec}. In this section, we focus on several of these problems. We run the proposed method on the datasets associated with each of these problems. However, instead of bypassing egomotion estimation as has been done in the original solutions to these problems, we use the egomotion computed from our algorithm for the tasks at hand. We show that for all the problems that we tested our approach on, the usage of our method results in either similar or improved performance.

We have implemented our version of \lsd-slam in C++, including short loop closures and the other heuristics mentioned in the last section. All the experiments have been carried out on a regular desktop with Core i7 2.3 GHz processor (containing 4 cores) and 16 GB RAM, running Ubuntu 14.04. Unoptimized implementation of the proposed algorithms runs at 2 to 3 frames per second on such a machine. We will release the source code of our implementation after the acceptance of the paper.

The proposed algorithm requires intrinsic parameters of the camera for egomotion estimation. For the sequences captured by us using GoPro Hero 3+ camera, we have calibrated the camera and used the estimated parameters. For the sequences taken from other sources, we do not have the information about the make and the version of the cameras used. Therefore for all these videos, we used the same set of calibration values as estimated by us for our GoPro Hero 3+. The approximate values appear to work reasonably well in our experiments, indicating the robustness of the proposed algorithm against camera calibration and distortion parameters. We would like to mention that we also tried autocalibration. However, owing to noisy and inaccurate depth estimates, the calibration estimation never converges. Most of the SLAM techniques that use self-calibration (like \cite{autocalib-cvpr16}) do so in controlled environments. In the absence of exact calibration, we have used standard parameters for GoPro cameras as mentioned earlier.

\begin{figure}[t]
\centering
\subfigure{\includegraphics[width=0.45\linewidth]{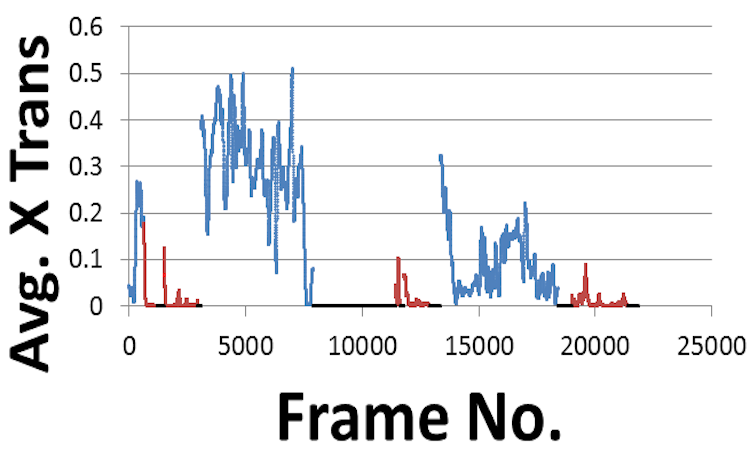} }
\subfigure{\includegraphics[width=0.45\linewidth]{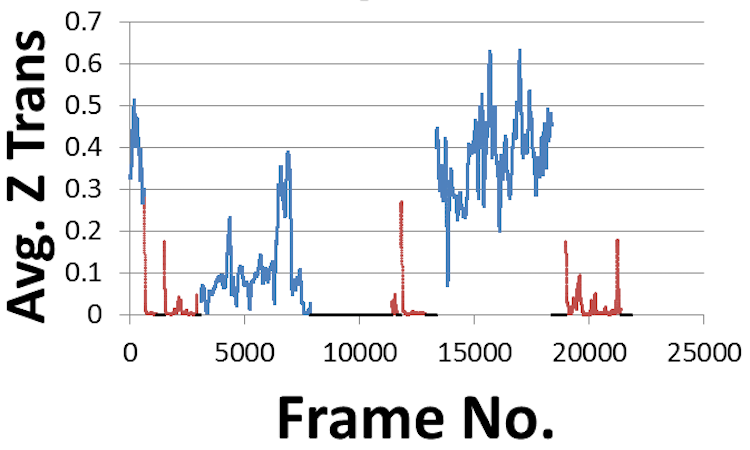} }
\caption{We show the egomotion computed using our technique on a video used in \cite{activity-rec,ego-seg}. The left and right figures show the computed X and Z translations respectively. The red curve in the plot shows the frames classified as stationary and the blue curve indicates transit frames. Black curve indicates unused frames in ours as well as the original paper}
\label{fig-exp-egoseg}
\end{figure}

In the egocentric videos, rotations form the major component of motion because of quick head movements (in comparison to typical slow forward translations). Rotation estimation accuracy is therefore a major requirement in such egomotion estimation problems. We experimented with rotation computations in both quaternion and Lie-algebraic frameworks. We found the Lie-algebra based computations to be more suitable as compared to the quaternion approach which often led to instabilities. The global loop closure techniques used in current state of the art, including LSD, ORB and DT SLAM, do not explicitly make use of the large number of redundant paths that are available in egocentric videos due to to-and-fro head movement. Also most of them rely on the noisy 3D estimates for loop closures. We would like to note that both the contributions (loop closure and rotation averaging using 2D estimates) are essential for the robust camera pose estimation. In our experiments, LSD SLAM with loop closures over short windows also fails due to the usage of noisy 3D estimates.

\begin{figure}[t]
\centering
\subfigure{ \includegraphics[width=0.45\linewidth]{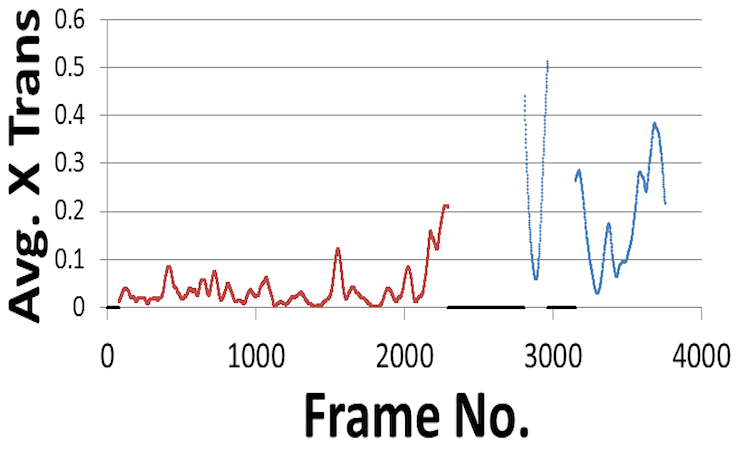} }
\subfigure{ \includegraphics[width=0.45\linewidth]{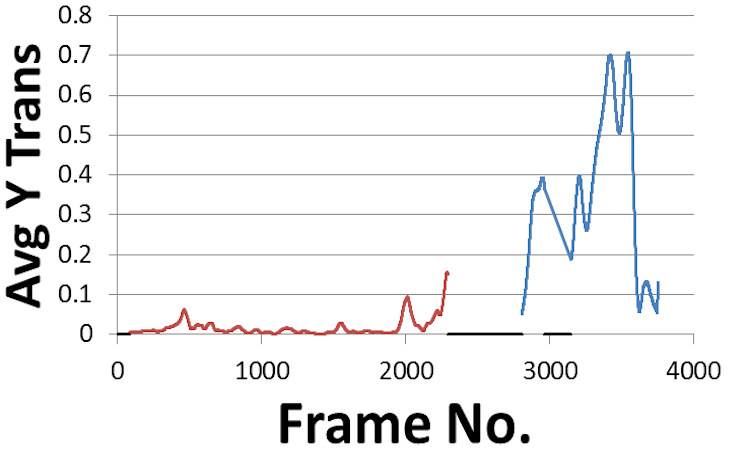} }
\caption{We use our algorithm to compute egomotion in a video from \cite{activity-rec} to detect `static' and `stair-climbing' activities.  Left and right images show X and Y translation components of the computed egomotion respectively. The red curve indicate frames classified as stationary, blue indicate stair-climbing and black show frames not used in the original as well as our classification.}
\label{Fig:tempseg}
\end{figure}

Since we do not have egocentric videos with ground truth poses (as described earlier, shaky handheld is not equivalent to egocentric), we have relied more on macro level accuracy cues such as overlaying the computed paths on maps (see Fig. \ref{Fig:trajectories}). We also synchronise the trajectories with the videos to manually verify if the significant events such as wearer turning are captured correctly. It may be noted that the observed instability of the LSD SLAM on the videos shown in Fig. \ref{Fig:trajectories} can not be attributed to calibration parameters as these were the sequences captured by us and exactly computed calibration parameters have been used for this experiment.

\subsection{Temporal Segmentation \cite{ego-seg}}

Poleg \etal \cite{ego-seg} have suggested a technique for temporally segmenting egocentric videos by labeling each frame as one of seven activities. The set of activities chosen by them forms a partition implying that at any point of time, a wearer must be in one of the seven activity states. Many of these activities such as walking, sitting etc. could have been classified by simple egomotion computation. However, they have reported the failure of pose computation on their videos and have suggested optic flow features as an alternative. To test the efficacy of our algorithm, we picked a random video sequence from their dataset, Huji\_Yair\_8\_part1. \lsd-slam fails on this sequence in consonance with the observation by Poleg \etal However our algorithm works flawlessly. We  classify frames into stationary and transit by computing inter-frame X and Z translations\footnote{In the 3D world coordinates, variations in depth, height and left \& right directions are marked on Z, Y and X axis respectively.} and accumulating them over windows of 10 frames. A SVM classifier is then trained over these features. We obtain an accuracy of 97\% for the two class classification problem against the reported accuracy of 98\% with the original flow based method.  Figure \ref{fig-exp-egoseg} shows the result.

\subsection{Activity Classification \cite{activity-rec}}

\begin{figure}[t]
\centering
\includegraphics[width=0.99\linewidth]{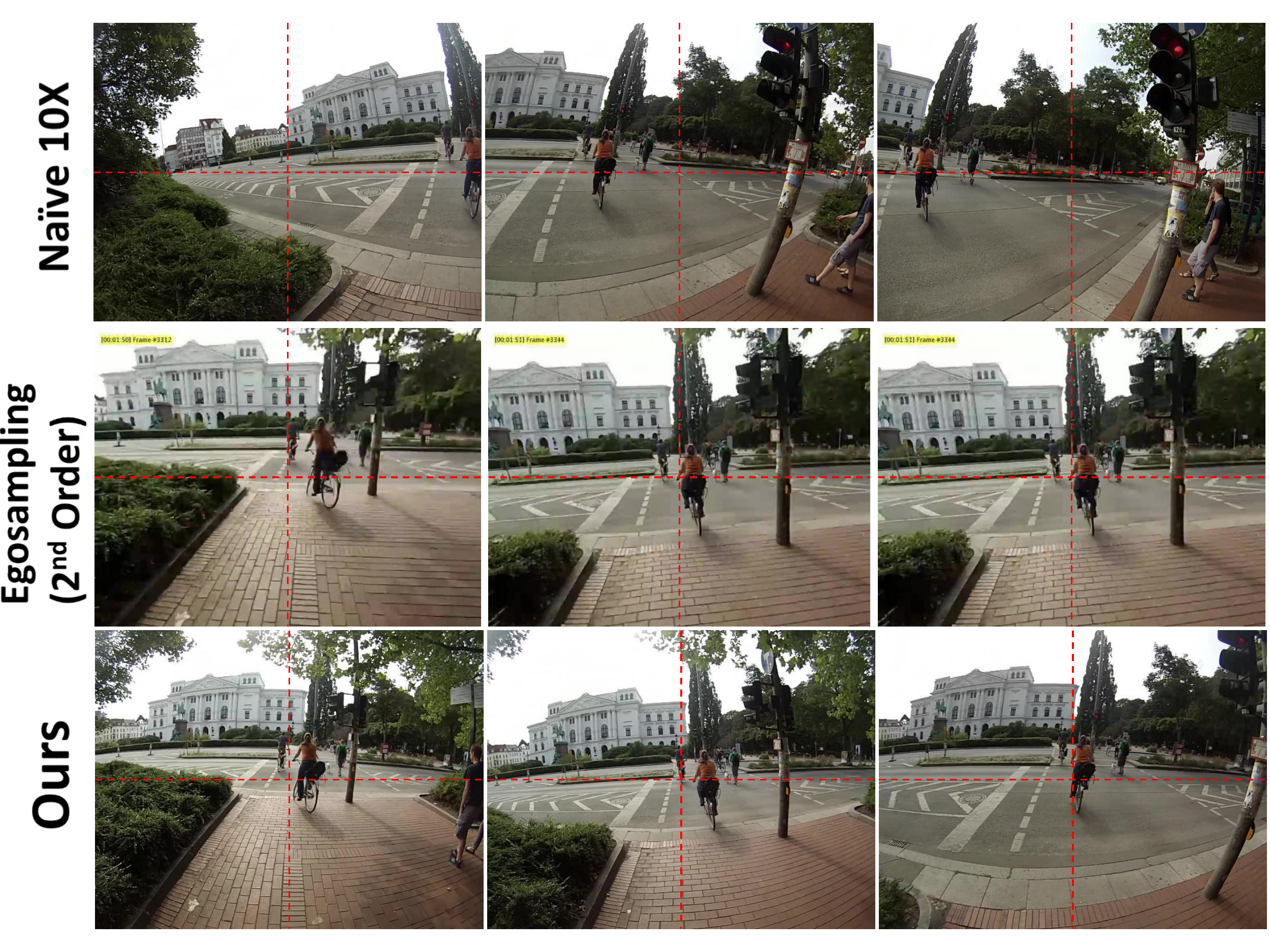}
\caption{Comparing proposed approach with naive $10\times$ fast forwarding and Egosampling \cite{ego-ff} on a publicly available video \cite{hyperlapse}. The first row shows output from uniform sampling. The second and third rows show outputs from EgoSampling and proposed approach. Focusing on the location of tree and the pedestrian reveals that both EgoSampling as well as proposed approach achieve equivalent stabilization which is much better compared to naive uniform sampling.}
\label{fig:egosample}
\end{figure}

In an another paper of theirs, Poleg \etal \cite{activity-rec} have extended the optical flow based technique presented in their original paper \cite{ego-seg}. Here, instead of temporal segmentation, they show how to train a compact CNN for classifying long term activities of the wearer based on the suggested sparse optical flow. While optical flow, can be computed reliably, it is an ambiguous cue for the considered egocentric activities. Camera egomotion is a much better indicator (if possible to compute reliably) and can improve the quality of prediction. However, as in their original work, Poleg \etal report the failure of SLAM techniques. To show the robustness of our algorithm, we used our algorithm to compute egomotion on one of their videos, ``2013 American Lung Association Fight For Air stair climb''. We then used the computed X and Y translation components of egomotion to distinguish between two classes, stationary vs stair climbing. Similar to the approach described in section 5.1, we accumulate these components over windows of 10 frames thereafter training a SVM classifier over it. We achieved an F1-score of 0.99 for both the classes whereas \cite{activity-rec} achieved an F1-score of 0.74 and 0.61 for stair climbing and static respectively. Figure~\ref{Fig:tempseg} shows the results. It may be noted that the purpose of the experiments is to show the usefulness of our technique for solving range of problems in variety of video capture settings. The selection of test sequence/classes is only indicative and random.

\begin{figure}[t]
\centering
\includegraphics[width=0.99\linewidth]{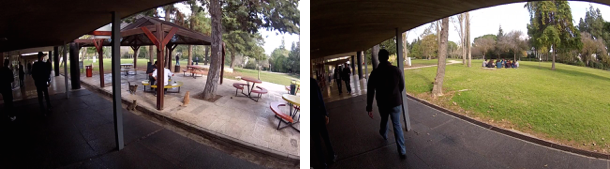}
\caption{Wearer's gaze fixation is easy to detect in an egocentric video by looking at the constancy of camera look at point. However, Poleg \etal reported failure of egomotion computation and suggested a flow based technique. We use the computed egomotion from our technique on their videos and successfully detect the gaze fixation instances. The figures show some fixations detected by our method.}
\label{fig:fixation}
\end{figure}

\begin{figure}[t]
\centering
\includegraphics[width=0.95\linewidth]{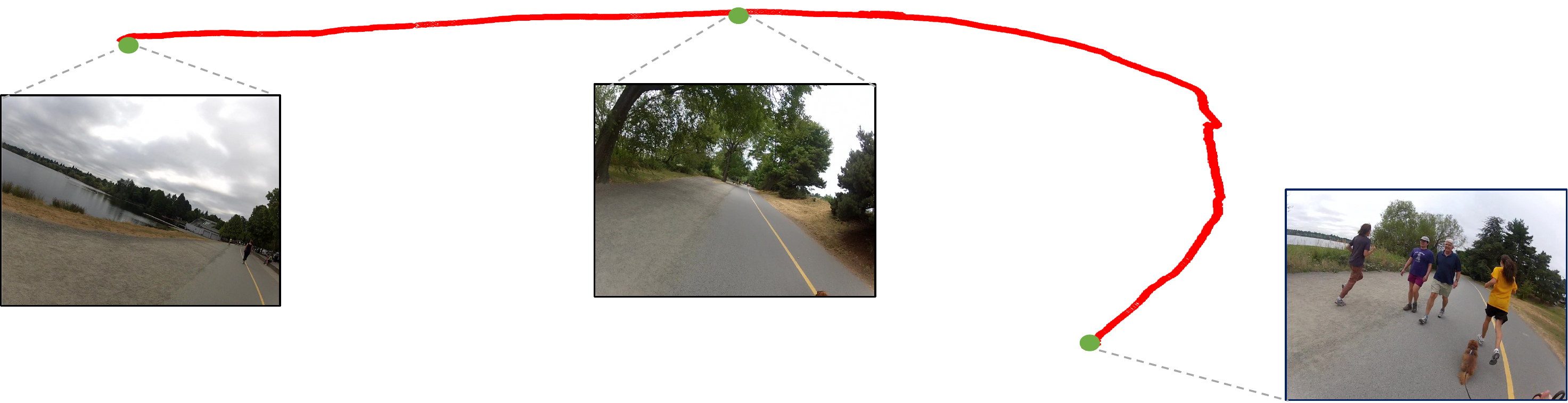}
\caption{We tested the proposed method on long shaky sequences given in Hyperlapse \cite{hyperlapse}. The trajectory shown is for video sequence gl02.mp4 in Hyperlapse dataset.}
\label{fig:hyperlapse}
\end{figure}

\begin{figure*}[t]
\centering
\subfigure[]{\includegraphics[height=0.16\linewidth]{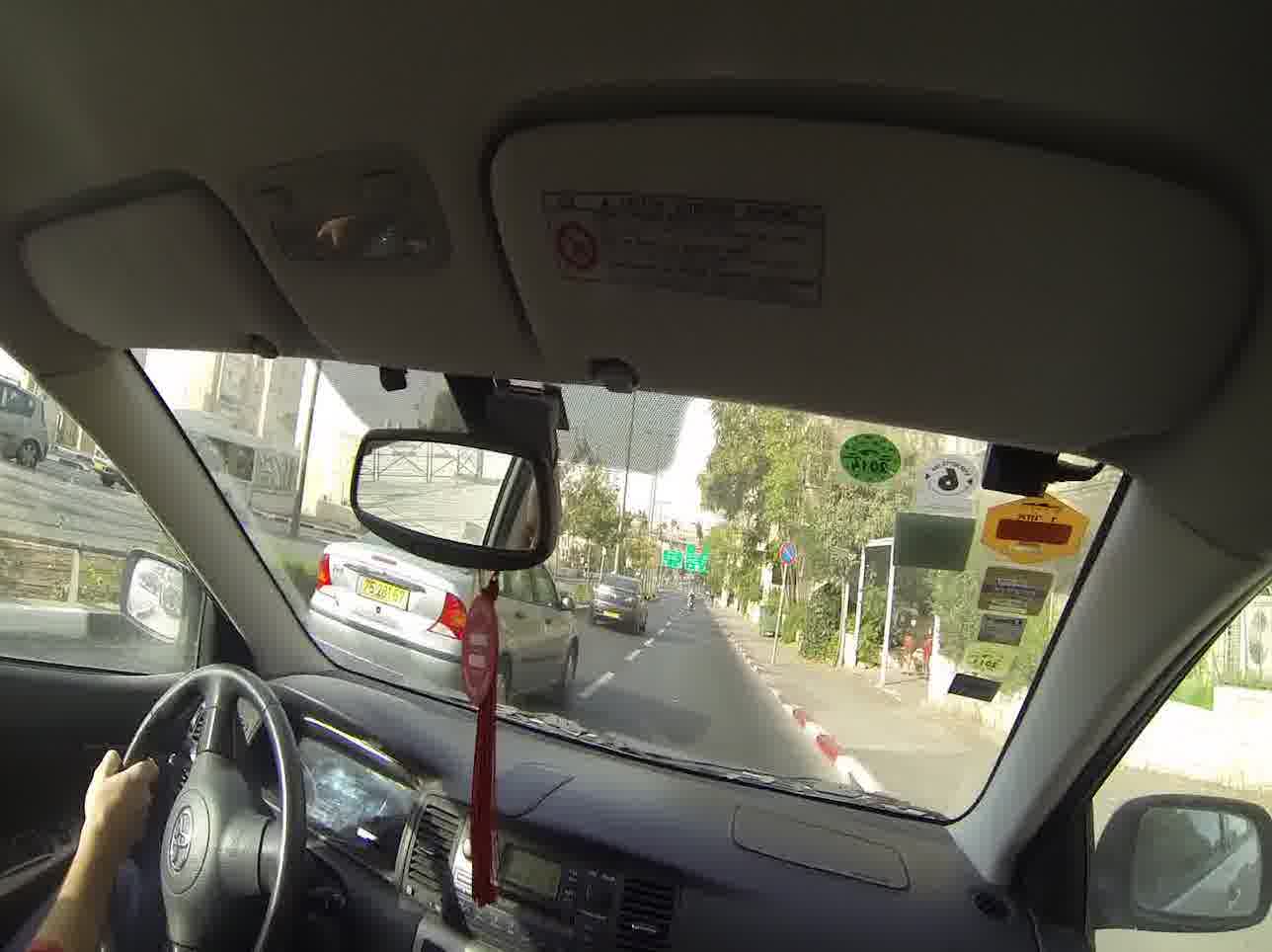}}\;
\subfigure[]{\includegraphics[height=0.16\linewidth]{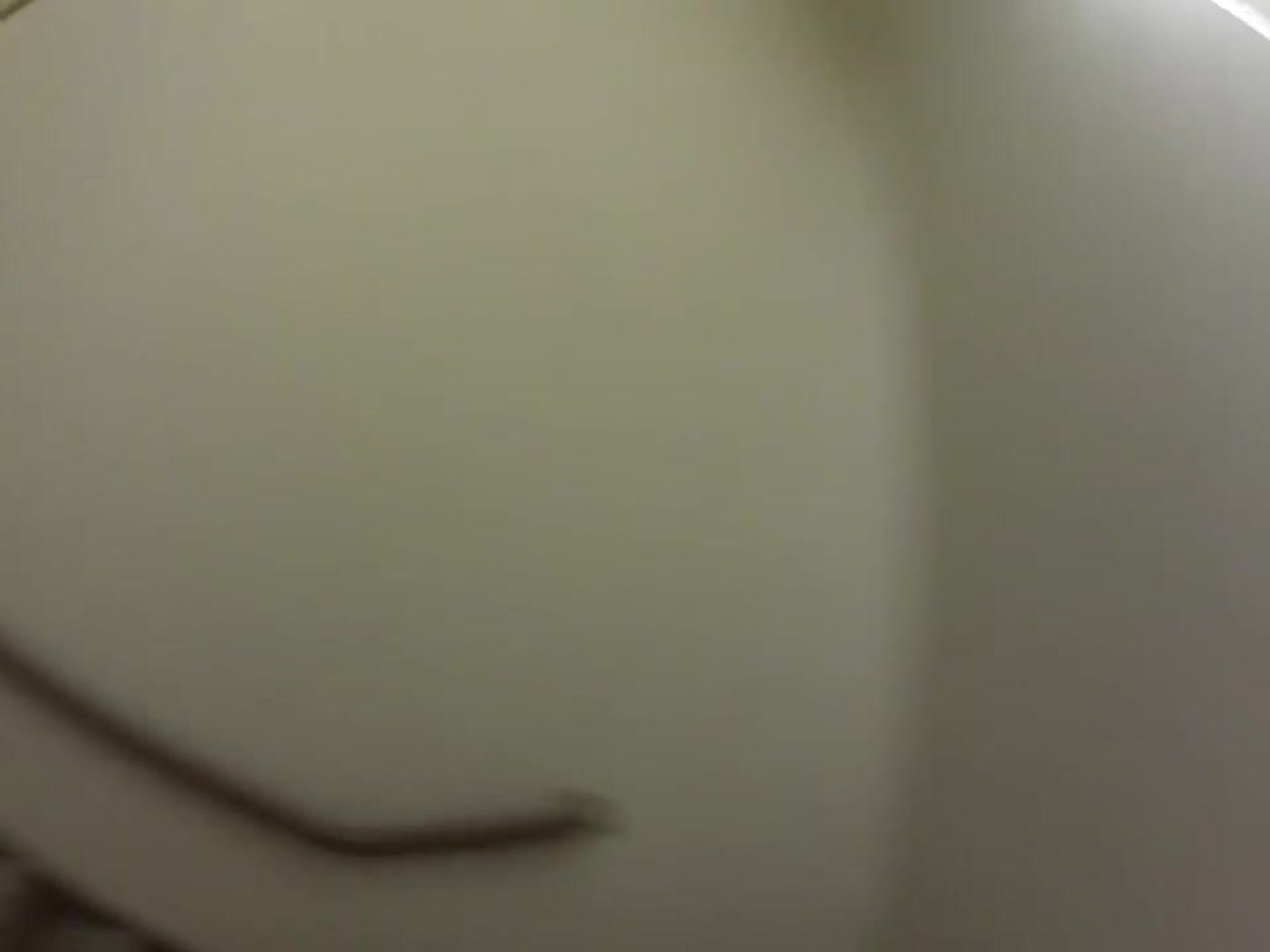}}\;
\subfigure[]{\includegraphics[height=0.16\linewidth]{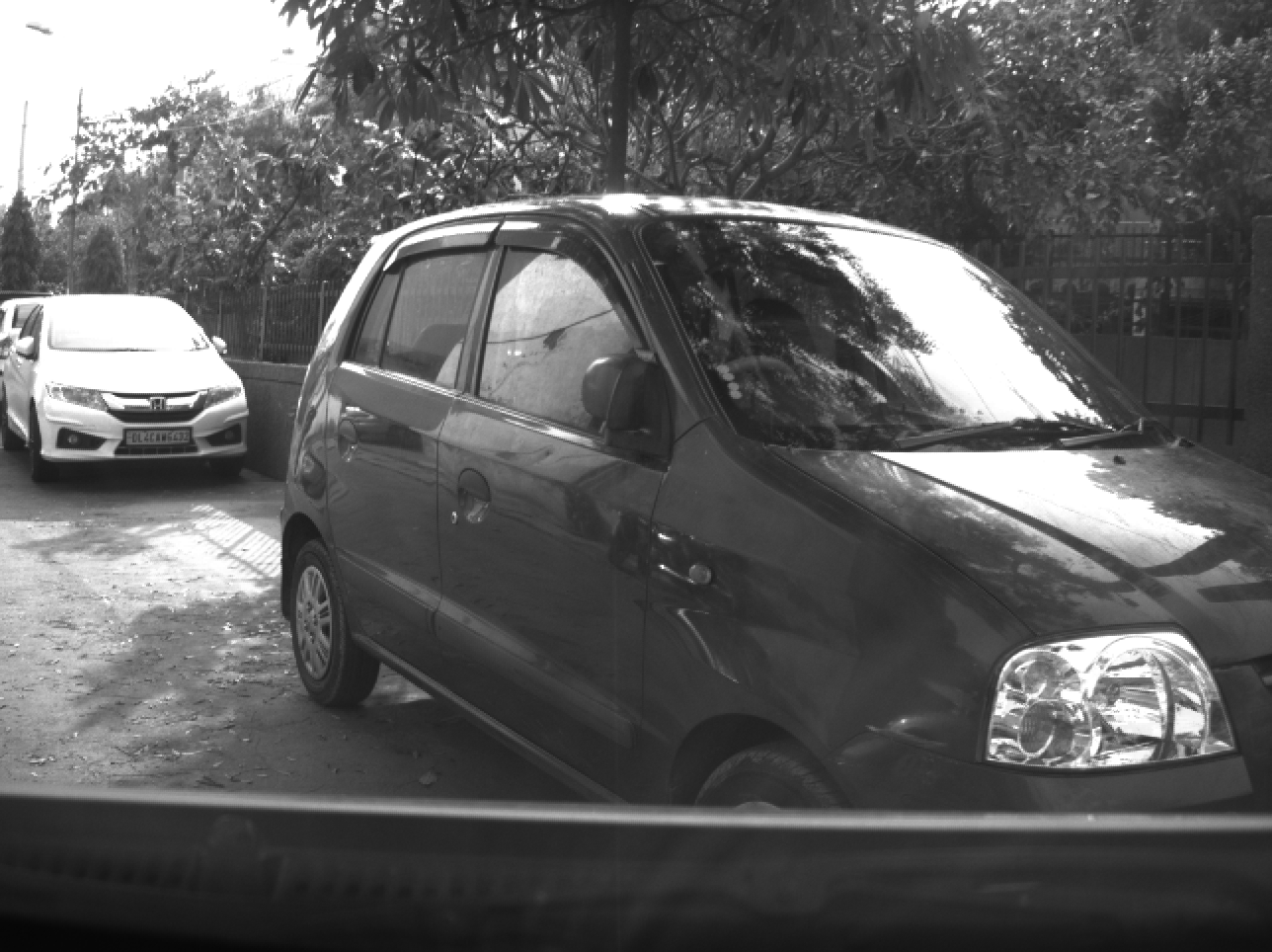}}\;
\subfigure[]{\includegraphics[height=0.16\linewidth]{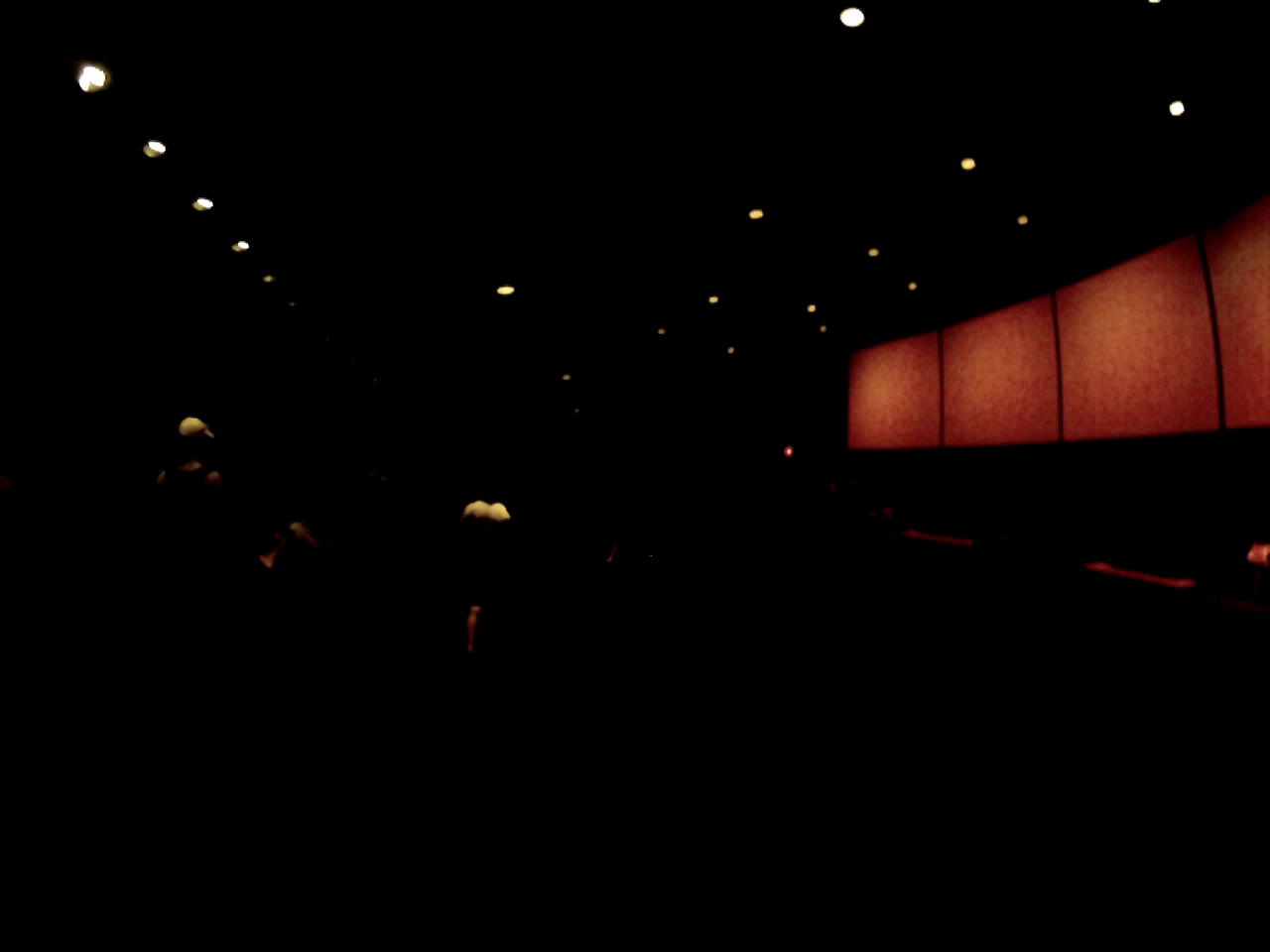}}
\caption{Our algorithm fails in the above cases. In (a) the wearer is part of multiple environments and (b) suffers from extreme blur. In (c) the scene is formed of non Lambertian surfaces causing inaccuracies in the pose estimates. (d) has negligible illumination}
\label{fig:failure-cases}
\end{figure*}

\subsection{Egosampling \cite{ego-ff}}

Egocentric videos tend to be long and shaky making them hard to watch. Fast forwarding is a natural solution to quickly browse such videos. However, fast forward by uniform sampling may accentuate the shake already present in egocentric videos. EgoSampling by Poleg  \etal is an adaptive frame sampling technique which models frame sampling and video stabilization as a joint optimization problem. The problem is posed as one of finding a shortest path in a graph where each node represents a frame in the video and the edge between frames $t$ and $k$ represents the perceived stability if frame $k$ is placed immediately after frame $t$ in the output. Amongst many costs, one of the costs suggested by Poleg \etal is the `shakiness' cost which biases the selection in favor of forward or similarly looking frames. They have reported failure of SLAM and have used focus of expansion calculated from optical flow for the purpose. We consider the same framework as EgoSampling \cite{ego-ff} but use our computed egomotion to compute the shakiness cost. Figure \ref{fig:egosample} shows that using our method one can achieve similar improvement in stabilization as EgoSampling compared to uniform sampling, but in a more principled manner.

\subsection{Gaze fixation \cite{ego-seg}}

Instead of continuous rotations of the camera due to the constant head movement of the wearer, egocentric videos sometime exhibit anomalies when this natural head rotation temporarily ceases. The anomalies usually indicate wearer's gaze fixation. Poleg \etal  have given a flow based technique to detect such gaze fixations \cite{ego-seg}. We use our technique to find the egomotion and estimate the camera view vector, corresponding to the direction of the wearer's gaze. Gaze fixation can then be detected by looking at the constancy of the camera look at point. As in the original paper, we use a moving window of 5 seconds to detect gaze fixation. Our hypothesis is based on identifying a point of intersection of the view vectors for the frames within this moving window.  We use the proposed approach to identify gaze fixation points in two videos from the ``HUJI EgoSeg dataset (Huji\_Chetan\_1 and Huji\_Yair\_2)''. The gaze annotation was not available in the dataset and we manually annotated the frames. From the 17 gaze fixation points, our method successfully detected 16 fixations with 3 false positives, which is comparable to what has been reported in the original work.

\subsection{Hyperlapse \cite{hyperlapse}}

Hyperlapse gives a technique to smoothly fast-forward a long hand-held or egocentric video with rapid camera motion \cite{hyperlapse}. The algorithm uses a structure from motion technique \cite{hyperlapse-sfm} to compute camera trajectory as well as the 3D map. Since the SfM algorithm do not scale well the videos were divided in batches of 1400 frames. We tested our algorithm as well as \lsd-slam on a video from their dataset. Figure \ref{fig:hyperlapse} shows the result. While \lsd-slam fails, our method works even on long sequences of 10,000 frames.

\subsection{Failure Cases}

Our experiments indicate that the proposed algorithm is much more robust compared to state of the art algorithms like \lsd-slam. However, we could identify multiple cases where the work still needs improvement. Figure \ref{fig:failure-cases} shows some failure cases. 

\section{Conclusion}

Egomotion is an important information in many egocentric applications. However, its use has been restricted until now because of failure of state-of-the-art visual SLAM techniques on such videos in the wild. We exploit the typical motion profile in an egocentric video to perform local loop closures based on realignment of wearer's gaze directions. This allows us to fix the camera pose locally and efficiently. Noting that camera motion is dominant 3D rotation, we use two step loop closure doing rotation averaging first followed by translation fixation. Our experiments indicate that exploiting these properties of egocentric videos leads to a robust camera pose estimation algorithm. Our experiments on many egocentric applications where egomotion has been reported to fail indicates that the proposed algorithm can be used successfully. This resolves a long standing problem in egocentric vision unlocking the use of egomotion in many egocentric applications.

{
\bibliographystyle{IEEEtran}
\bibliography{egbib}

\begin{thebibliography}{10}
\providecommand{\url}[1]{#1}
\csname url@samestyle\endcsname
\providecommand{\newblock}{\relax}
\providecommand{\bibinfo}[2]{#2}
\providecommand{\BIBentrySTDinterwordspacing}{\spaceskip=0pt\relax}
\providecommand{\BIBentryALTinterwordstretchfactor}{4}
\providecommand{\BIBentryALTinterwordspacing}{\spaceskip=\fontdimen2\font plus
\BIBentryALTinterwordstretchfactor\fontdimen3\font minus
  \fontdimen4\font\relax}
\providecommand{\BIBforeignlanguage}[2]{{%
\expandafter\ifx\csname l@#1\endcsname\relax
\typeout{** WARNING: IEEEtran.bst: No hyphenation pattern has been}%
\typeout{** loaded for the language `#1'. Using the pattern for}%
\typeout{** the default language instead.}%
\else
\language=\csname l@#1\endcsname
\fi
#2}}
\providecommand{\BIBdecl}{\relax}
\BIBdecl

\bibitem{ego-ff}
Y.~Poleg, T.~Halperin, C.~Arora, and S.~Peleg, ``{EgoSampling}: Fast-forward
  and stereo for egocentric videos,'' in \emph{Proceedings of the IEEE
  Conference on Computer Vision and Pattern Recognition (CVPR)}, 2015, pp.
  4768--4776.

\bibitem{hyperlapse}
J.~Kopf, M.~F. Cohen, and R.~Szeliski, ``First-person hyper-lapse videos,''
  \emph{ACM Transactions on Graphics (TOG)}, vol.~33, no.~4, pp. 78:1--78:10,
  2014.

\bibitem{ego-seg}
Y.~Poleg, C.~Arora, and S.~Peleg, ``Temporal segmentation of egocentric
  videos,'' in \emph{Proceedings of the IEEE Conference on Computer Vision and
  Pattern Recognition (CVPR)}, 2014, pp. 2537--2544.

\bibitem{activity-rec}
Y.~Poleg, A.~Ephrat, S.~Peleg, and C.~Arora, ``Compact {CNN} for indexing
  egocentric videos,'' in \emph{Proceedings of the IEEE Winter Conference on
  Applications of Computer Vision (WACV)}, 2016.

\bibitem{lsd-slam}
J.~Engel, T.~Schops, and D.~Cremers, ``{LSD-SLAM: Large-Scale Direct Monocular
  SLAM},'' in \emph{Proceedings of the European Conference on Computer Vision
  (ECCV)}, 2014, pp. 834--849.

\bibitem{google-glass}
\BIBentryALTinterwordspacing
Google, ``Glass,'' https://www.google.com/glass/start. [Online]. Available:
  \url{https://www.google.com/glass/start}
\BIBentrySTDinterwordspacing

\bibitem{gopro}
\BIBentryALTinterwordspacing
{GoPro}, ``Hero3,'' https://gopro.com/. [Online]. Available:
  \url{https://gopro.com/}
\BIBentrySTDinterwordspacing

\bibitem{ego_ac_recog_1}
T.~Starner, J.~Weaver, and A.~Pentland, ``{Real-Time American Sign Language
  Recognition Using Desk and Wearable Computer Based Video},'' \emph{IEEE
  Transactions on Pattern Analysis and Machine Intelligence (PAMI)}, vol.~20,
  no.~12, pp. 1371--1375, 1998.

\bibitem{ego_ac_recog_2}
H.~Pirsiavash and D.~Ramanan, ``Detecting activities of daily living in
  first-person camera views,'' in \emph{Proceedings of the IEEE Conference on
  Computer Vision and Pattern Recognition (CVPR)}, 2012, pp. 2847--2854.

\bibitem{ego_ac_recog_3}
A.~Fathi, A.~Farhadi, and J.~M. Rehg, ``Understanding egocentric activities,''
  in \emph{Proceedings of the IEEE International Conference on Computer Vision
  (ICCV)}, 2011, pp. 407--414.

\bibitem{ego_ac_regog_5_lowres}
S.~Sundaram and W.~Mayol-Cuevas, ``High level activity recognition using low
  resolution wearable vision,'' in \emph{Proceedings of the IEEE Computer
  Vision and Pattern Recognition Workshops (CVPRW)}, 2009, pp. 25--32.

\bibitem{jpl}
M.~S. Ryoo and L.~Matthies, ``First-person activity recognition: What are they
  doing to me?'' in \emph{Proceedings of the IEEE Conference on Computer Vision
  and Pattern Recognition (CVPR)}, 2013, pp. 2730--2737.

\bibitem{Grauman-Story:CVPR13}
Z.~Lu and K.~Grauman, ``Story-driven summarization for egocentric video,'' in
  \emph{Proceedings of the IEEE Conference on Computer Vision and Pattern
  Recognition (CVPR)}, 2013, pp. 2714--2721.

\bibitem{ego_ac_regco_and_chaptering}
K.~M. Kitani, T.~Okabe, Y.~Sato, and A.~Sugimoto, ``Fast unsupervised
  ego-action learning for first-person sports videos,'' in \emph{Proceedings of
  the IEEE Conference on Computer Vision and Pattern Recognition (CVPR)}, 2011,
  pp. 3241--3248.

\bibitem{Grauman-Important:CVPR2012}
Y.~J. Lee, J.~Ghosh, and K.~Grauman, ``Discovering important people and objects
  for egocentric video summarization,'' in \emph{Proceedings of the IEEE
  Conference on Computer Vision and Pattern Recognition (CVPR)}, 2012, pp.
  1346--1353.

\bibitem{rgbd-vo-icra}
C.~Kerl, J.~Sturm, and D.~Cremers, ``Robust odometry estimation for {RGB-D}
  cameras.'' in \emph{Proceedings of the IEEE International Conference on
  Robotics and Automation (ICRA)}, 2013, pp. 3748--3754.

\bibitem{semi-dense-vo-iccv-13}
J.~Engel, J.~Sturm, and D.~Cremers, ``Semi-dense visual odometry for a
  monocular camera,'' in \emph{Proceedings of the IEEE International Conference
  on Computer Vision (ICCV)}, 2013, pp. 1449--1456.

\bibitem{ptam}
G.~Klein and D.~Murray, ``Parallel tracking and mapping on a camera phone,'' in
  \emph{Proceedings of the {IEEE} International Symposium on Mixed and
  Augmented Reality {(ISMAR)}}, 2009, pp. 83--86.

\bibitem{dtam}
R.~A. Newcombe, S.~J. Lovegrove, and A.~J. Davison, ``{DTAM}: Dense tracking
  and mapping in real-time,'' in \emph{Proceedings of the IEEE International
  Conference on Computer Vision (ICCV)}, 2011, pp. 2320--2327.

\bibitem{govinduefficient}
A.~Chatterjee and V.~M. Govindu, ``Efficient and robust large-scale rotation
  averaging.'' in \emph{Proceedings of the IEEE International Conference on
  Computer Vision (ICCV)}, 2013, pp. 521--528.

\bibitem{orb-slam}
R.~Mur{-}Artal, J.~M.~M. Montiel, and J.~D. Tard{\'{o}}s, ``{ORB-SLAM:} a
  versatile and accurate monocular {SLAM} system,'' \emph{IEEE Transactions on
  Robotics}, vol.~31, no.~5, pp. 1147--1163, 2015.

\bibitem{Williams2009}
B.~Williams, M.~Cummins, J.~Neira, P.~Newman, I.~Reid, and J.~Tard\'{o}s, ``A
  comparison of loop closing techniques in monocular {SLAM},'' \emph{Robotics
  and Autonomous Systems}, vol.~57, no.~12, pp. 1188--1197, 2009.

\bibitem{clemente_etal_rss2007}
L.~Clemente, A.~Davison, I.~Reid, J.~Neira, and J.~D. Tard\'{o}s, ``Mapping
  large loops with a single hand-held camera,'' in \emph{Proceedings of
  Robotics: Science and Systems Conference}, 2007.

\bibitem{CumminsIJRR08}
M.~Cummins and P.~Newman, ``{FAB-MAP: Probabilistic Localization and Mapping in
  the Space of Appearance},'' \emph{The International Journal of Robotics
  Research}, vol.~27, no.~6, pp. 647--665, 2008.

\bibitem{WilliamsIROS08}
B.~Williams, M.~Cummins, J.~Neira, P.~Newman, I.~Reid, and J.~D. Tardos, ``An
  image-to-map loop closing method for monocular {SLAM},'' in \emph{Proceeding
  of the International Conference on Intelligent Robots and Systems}, 2008, pp.
  2053--2059.

\bibitem{homo}
M.~J. Leotta, E.~Smith, M.~Dawkins, and P.~Tunison, ``Open source
  structure-from-motion for aerial video,'' in \emph{2016 IEEE Winter
  Conference on Applications of Computer Vision (WACV)}, 2016, pp. 1--9.

\bibitem{KLDiv}
P.~J. Moreno, P.~Ho, and N.~Vasconcelos, ``A {Kullback-Leibler} divergence
  based kernel for svm classification in multimedia applications.'' in
  \emph{Proceedings of Advances in Neural Information Processing Systems
  (NIPS)}, 2004, pp. 1385--1392.

\bibitem{social-interaction}
A.~Fathi, J.~K. Hodgins, and J.~M. Rehg, ``Social interactions: A first-person
  perspective,'' in \emph{Proceedings of the IEEE Conference on Computer Vision
  and Pattern Recognition (CVPR)}, 2012, pp. 1226--1233.

\bibitem{autocalib-cvpr16}
H.~Ha, S.~Im, J.~Park, H.-G. Jeon, and I.~S. Kweon, ``High-quality depth from
  uncalibrated small motion clip,'' in \emph{Proceedings of IEEE Conference on
  Computer Vision and Pattern Recognition (CVPR)}, 2016.

\bibitem{hyperlapse-sfm}
C.~Wu, ``Towards linear-time incremental structure from motion,'' in
  \emph{Proceedings of the International Conference on 3D Vision (3DV)}, 2013,
  pp. 127--134.

\end{thebibliography}
}

\end{document}